\documentclass{article}

\usepackage{microtype}
\usepackage{graphicx}
\usepackage{subfigure}
\usepackage{booktabs} 
\usepackage{textcomp}
\usepackage{hyperref}

\usepackage[accepted]{sysml2019}

\usepackage[english]{babel}
\usepackage[utf8x]{inputenc}
\usepackage[T1]{fontenc}

\usepackage{amsmath}
\usepackage[colorinlistoftodos]{todonotes}

\newif\ifnote
\notetrue

\newif\iffbone
\fbonetrue

\begin{document}
\twocolumn[
\sysmltitle{PyTorch-BigGraph: A Large-scale Graph Embedding System}

\begin{sysmlauthorlist}
\sysmlauthor{Adam Lerer}{fb}
\sysmlauthor{Ledell Wu}{fb}
\sysmlauthor{Jiajun Shen}{fb}
\sysmlauthor{Timothee Lacroix}{fb}
\sysmlauthor{Luca Wehrstedt}{fb}
\sysmlauthor{Abhijit Bose}{fb}
\sysmlauthor{Alex Peysakhovich}{fb}
\end{sysmlauthorlist}
\sysmlaffiliation{fb}{Facebook AI Research, New York, NY, USA}

\sysmlcorrespondingauthor{Adam Lerer}{alerer@fb.com}

\sysmlkeywords{graph embeddings,representation learning}

\vskip 0.3in

\begin{abstract}

Graph embedding methods produce unsupervised node features from graphs that can then be used for a variety of machine learning tasks. Modern graphs, particularly in industrial applications, contain billions of nodes and trillions of edges, which exceeds the capability of existing embedding systems. We present PyTorch-BigGraph (PBG), an embedding system that incorporates several modifications to traditional multi-relation embedding systems that allow it to scale to graphs with billions of nodes and trillions of edges. PBG uses graph partitioning to train arbitrarily large embeddings on either a single machine or in a distributed environment. We demonstrate comparable performance with existing embedding systems on common benchmarks, while allowing for scaling to arbitrarily large graphs and parallelization on multiple machines. We train and evaluate embeddings on several large social network graphs as well as the full Freebase dataset, which contains over 100 million nodes and 2 billion edges.
\end{abstract}
]

\printAffiliationsAndNotice{}  

\section{Introduction}


Graph structured data is a common input to a variety of machine learning tasks \cite{wu2017starspace,cook2006mining,Nickel_review_2016,hamilton2017representation}. Working with graph data directly is difficult, so a common technique is to use graph embedding methods to create vector representations for each node so that distances between these vectors predict the occurrence of edges in the graph.  Graph embeddings have been have been shown to serve as useful features for downstream tasks such as recommender systems in e-commerce \cite{wang2018billion}, link prediction in social media \cite{deepwalk14}, predicting drug interactions and characterizing protein-protein networks \cite{ZitnikL17}. 



Graph data is common at modern web companies and poses an extra challenge to standard embedding methods: scale. For example, the Facebook graph includes over two billion user nodes and over a trillion edges representing friendships, likes, posts and other connections \cite{ching2015one}. The graph of users and products at Alibaba also consists of more than one billion users and two billion items \cite{wang2018billion}. At Pinterest, the user to item graph includes over 2 billion entities and over 17 billion edges \cite{ying2018graph}.

There are two main challenges for embedding graphs of this size. First, an embedding system must be fast enough to embed graphs with $10^{11} - 10^{12}$ edges in a reasonable time. Second, a model with two billion nodes and 100 embedding parameters per node (expressed as floats) would require 800GB of memory just to store its parameters, thus many standard methods exceed the memory capacity of typical commodity servers.

We present PyTorch-BigGraph (PBG), an embedding system that incorporates several modifications to standard models. The contribution of PBG is to scale to graphs with billions of nodes and trillions of edges. Important components of PBG are:

\begin{itemize}
\item A block decomposition of the adjacency matrix into $N$ buckets, training on the edges from one bucket at a time. PBG then either swaps embeddings from each partition to disk to reduce memory usage, or performs distributed execution across multiple machines.
\item A distributed execution model that leverages the block decomposition for the large parameter matrices, as well as a parameter server architecture for global parameters and feature embeddings for featurized nodes.
\item Efficient negative sampling for nodes that samples negative nodes both uniformly and from the data, and reuses negatives within a batch to reduce memory bandwidth.
\item Support for multi-entity, multi-relation graphs with per-relation configuration options such as edge weight and choice of relation operator.
\end{itemize}

We evaluate PBG on the Freebase, LiveJournal and YouTube graphs and show that it matches the performance of existing embedding systems.

We also report results on larger graphs. We construct an embedding of the full Freebase knowledge graph (121 million entities, 2.4 billion edges), which we release publicly with this paper. Partitioning of the Freebase graph reduces memory consumption by 88\% with only a small degradation in the embedding quality,  and distributed execution on 8 machines decreases training time by a factor of 4. We also perform experiments on a large Twitter graph showing similar results with near-linear scaling.

PBG has been released as an open source project at \url{https://github.com/facebookresearch/PyTorch-BigGraph}. It is written entirely in Pytorch \cite{paszke2017automatic} with no external dependencies or custom operators.

\section{Related Work}
Many types of models have been developed for multi-relation graphs \cite{bordes2011learning,transE,nickel2011three,trouillon2016complex}. Typically these models  have been used in the context of entity representations in knowledge bases (e.g. Freebase or WordNet). Entities are given a base vector, these vectors are transformed by a learned function for each transformation, and existence of edges is predicted by some distance measure in the new space. More recent work by \citeauthor{wu2017starspace} proposes modeling some entities as bags of other entities (rather than giving them explicit embeddings). PBG borrows many insights on loss functions and transformations from this literature.

There are significant engineering challenges to scaling graph embedding models. Proposed approaches in the literature include multi-level methods \cite{liang2018mile}, distributed embedding systems \cite{gridword2vec, Swivel2016}, as well as specialized methods for standard algorithms such as SVD and k-means on large graphs \cite{ching2015one}. Gains from large embedding systems have been documented in e-commerce \cite{wang2018billion} and other applications. 

There is an extensive literature on distributional semantics in natural language processing. A key breakthrough in this literature are algorithms such as word2vec which allowed word embedding methods to scale to larger corpora \cite{mikolov2013efficient}. Recent work has shown that there is economic value from ingesting even larger data sets using distributed word2vec systems \cite{gridword2vec}. 

There is substantial prior work on scalable parallel algorithms for training machine learning models \cite{Dean2012}. Highly related to PBG is work on scaling various forms of matrix factorization \cite{gupta1997highly,Gemulla:2011:LMF:2020408.2020426}. Matrix factorization is closely related to embeddings, and has had widespread success in recommender systems \cite{koren2009matrix}.

Recent work proposes to construct embeddings by using graph convolutional neural networks (GCNs, \citealt{kipf2016semi}). These methods have shown success when applied to problems at large-scale web companies \cite{hamilton2017inductive,ying2018graph}. The problem studied by the GCN is different than the one solved by PBG (mostly in that GCNs are typically applied to graphs where the nodes are already featurized). Combining ideas from graph embedding and GCN models is an interesting future direction both for theory and applications.

\begin{figure*}[t]
\centering
\includegraphics[width=12cm]{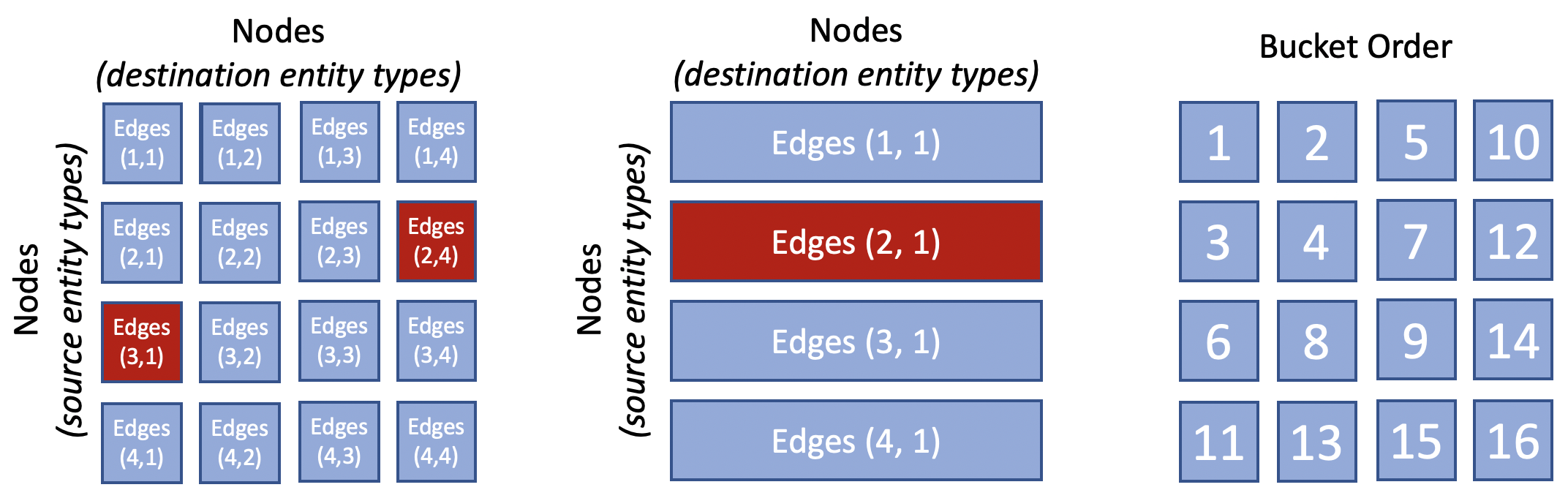}
\caption{The PBG partitioning scheme for large graphs. \textbf{Left:} nodes are divided into $P$ partitions that are sized to fit in memory. Edges are divided into buckets based on the partition of their source and destination nodes. In distributed mode, multiple buckets with non-overlapping partitions can be executed in parallel (red squares). \textbf{Center:} Entity types with small cardinality do not have to be partitioned; if all entity types used for tail nodes are unpartitioned, then edges can be divided into $P$ buckets based only on source node partitions. \textbf{Right:} the `inside-out' bucket order guarantees that buckets have at least one previously-trained embedding partition. Empirically, this ordering produces better embeddings than other alternatives (or random)}
\label{fig:partitions}
\end{figure*}

\section{Multi-Relation Embeddings}
\subsection{Model}
\label{section:model_description}

A multi-relation graph is a directed graph $G=(V,R,E)$ where $V$ are the nodes (aka entities), $R$ is a set of relations, and $E$ is a set of edges where a generic element $e = (s, r, d)$ (source, relation, destination) where $s, d\in V$ and $r \in R.$ We also discuss graphs that have multiple \textit{entity types}. Such graphs have a set of entity types and a mapping from nodes to entity types, and each relation specifies a single entity type for source and destination nodes for all edges of that relation.

We will represent each entity and relation type with a vector of parameters. We will denote this vector as $\theta$. A multi-relation graph embedding uses a score function $f(\theta_s, \theta_r, \theta_d)$ that produces a score for each edge that attempts to maximize the score of $f(\theta_s, \theta_r, \theta_d)$ for any $(s, r, d) \in E$ and minimizes it for $(s, r, d) \not \in E.$

PBG considers scoring functions between a transformed version of an edge's source and destination entities' vectors ($\theta_s, \theta_d$): $$f(\theta_s, \theta_r, \theta_d)=sim\left(g_{(s)}(\theta_s, \theta_r), g_{(d)}(\theta_d, \theta_r)\right)$$ where $\theta_r$ corresponds to parameters of the relation-specific transformation operator. Using a factorized scoring function produces a embeddings where the (transformed) similarity between node embeddings has semantic meaning.

PBG uses dot product or cosine similarity scoring functions, and a choice of relation operator $g$ which include linear transformation, translation, and complex multiplication. This combination of scoring functions and relation operators allows PBG to train RESCAL, DistMult, TransE, and ComplEx models \cite{nickel2011three,yang2014embedding,transE,trouillon2016complex}. \footnote{For knowledge base datasets, state-of-the-art performance is achieved with ComplEx embeddings, but this may not generalize to all graphs. On small knowledge graphs, a general linear transform (RESCAL) does not perform as well as transformations with fewer parameters such as translation (as well as transformations that can be represented in the RESCAL model) because the relation operators overfit \cite{hole}. However, we are interested in web interaction graphs which have a very small number of relations relative to entities, so the relation parameters do not contribute substantially to model size, nor are they prone to overfitting.} A subset of relation types may use the identity relation, so that the untransformed entity embeddings predict edges of this relation.

\begin{table}[ht]
\centering
\begin{tabular}{| l | c c | }
\hline
\textbf{Model} & ${\bf g(x,\theta_r)}$ & ${\bf sim(a, b)}$\\
\hline
RESCAL & $A_r x$ & $<a, b>$\\
TransE & $x + \theta_r$ & $cos(a, b)$\\
DistMult & $x \odot\ \theta_r$ & $<a, b>$ \\
ComplEx & $x \odot\ \theta_r$ & $Re\{<a, \overline{b}>\}$ \\
\hline 
\end{tabular}
\end{table}

We consider sparse graphs, so the input to PBG is a list of positive-labeled (existing) edges. Negative edges are constructed by sampling. In PBG negative samples are generated by corrupting positive edges by sampling either a new source or a destination for each existing edge \cite{transE}.

Because edge distributions in real world graphs are heavy tailed, the choice of how to sample nodes to construct negative examples can affect model quality \cite{mikolov2013efficient}. On one hand, if we sample negatives strictly according to the data distribution, there is no penalty for the model predicting high scores for edges with rare nodes. On the other hand, if we sample negatives uniformly, the model can perform very well (especially in large graphs) by simply scoring edges proportional to their source and destination node frequency in the dataset. Both of these results are undesirable, so in PBG we sample a fraction $\alpha$ of negatives according to their prevalence in the training data and $(1-\alpha)$ of them uniformly at random. By default PBG uses $\alpha=.5.$ 

In multi-entity graphs, negatives are only sampled from the correct entity type for an edge's relation. Thus, in our model, the score for an `invalid' edge (wrong entity types) is undefined. The approach of using entity types has been studied before in the context of knowledge graphs \cite{krompass2015type}, but we found it to be particularly important in graphs that have entity types with highly unbalanced numbers of nodes, e.g. 1 billion users vs. 1 million products. With uniform negative sampling over all nodes, the loss would be dominated by user negative nodes and would not optimize for ranking between user-product edges.

PBG optimizes a margin-based ranking objective between each edge $e$ in the training data and a set of edges $e'$ constructed by corrupting $e$ with either a sampled source or destination node (but not both).

\[
\label{math:loss}
\mathcal{L} = \sum_{\substack{e \in G}} \sum_{e'\in S'_e} \max(f(e) - f(e') + \lambda, 0))
\]

where $\lambda$ is a margin hyperparameter and

\[
\label{math:S}
S'_e = \left\{(s',r,d)|s'\in V\right\} \cup \left\{(s,r,d'|d'\in V\right\}.
\]

Logistic and softmax loss functions may also be used instead of a ranking objective in order to reproduce certain graph embedding models (e.g. \citealt{trouillon2016complex}).

Model updates to the embeddings and relation parameters are performed via minibatch stochastic gradient descent (SGD). We use the Adagrad optimizer, and sum the accumulated gradient $\mathcal{G}$ over each embedding vector to reduce memory usage on large graphs \cite{adagrad}.

\section{Training at Scale}
\label{sec:large-scale}

PBG is designed to operate on arbitrarily large graphs running on either a single machine or can be distributed across multiple machines. In either case, training occurs on a number of CPU threads equal to the number of machine cores, with no explicit synchronization between cores as described in \cite{hogwild}. 


\subsection{Partitioning of Entities and Edges}
\label{sec:partition}
PBG uses a partitioning scheme to support models that are too large to fit in memory on a single machine. This partitioning also allows for distributed training of the model. 

Each entity type in $G$ can be either partitioned or remain unpartitioned. Partitioned entities are split into $P$ parts. $P$ is chosen such that each part fits into memory or to support the desired level of parallelism for execution.

After entities are partitioned, edges are divided into \textit{buckets} based on their source and destination entities' partitions. For example, if an edge has a source in partition $p_1$ and destination in partition $p_2$ then it is placed into bucket $(p_1, p_2).$ This creates $P^2$ buckets when both source and destination entity types are partitioned and $P$ buckets if only source (or destination) entities are partitioned.

Each epoch of training iterates through each of the edge buckets. For edge bucket $(p_i, p_j)$, source and destination partitions $i$ and $j$ respectively are swapped from disk, and then the edges (or a subset of edges) are loaded and subdivided among the threads for training.

This graph partitioning introduces two functional changes to the base algorithm described in the last section. First, each candidate edge $(s,r,d)$ is only compared to negatives $(s,r,d')$ in the ranking loss where $d'$ is drawn from the same partition (same for source nodes)\footnote{This would not matter if we were using an independent loss for positives and negatives, e.g. a binary cross-entropy loss}. 

Second, edges are no longer sampled i.i.d. but are grouped by partition. Convergence under SGD to a stationary or chain-recurrent point, is still guaranteed under this modification (see \cite{Gemulla:2011:LMF:2020408.2020426}, Sec. 4.2), but may suffer from slower convergence\footnote{The slower convergence may be ameliorated by switching between the buckets (`stratum losses' \cite{Gemulla:2011:LMF:2020408.2020426}) more frequently, i.e. in each epoch divide the edges from each bucket into $N$ parts and iterate over the buckets $N$ times, operating on one edge part each time.}\footnote{In practice, we use Adagrad rather than SGD.}.

We observe that the order of iterating through edge buckets may affect the final model. Specifically, for each edge bucket $(p_1,p_2)$ except the first, it is important that an edge bucket $(p_1, *)$ or $(*, p_2)$ was trained in a previous iteration. This constraint ensures that embeddings in all partitions are aligned in the same space. For single-machine embeddings, we found that an `inside-out` ordering, illustrated in Figure \ref{fig:partitions}, achieved the best performance while minimizing the number of swaps to disk. 

\subsection{Distributed Execution}

\begin{figure*}[t]
\centering
\includegraphics[width=0.8\textwidth]{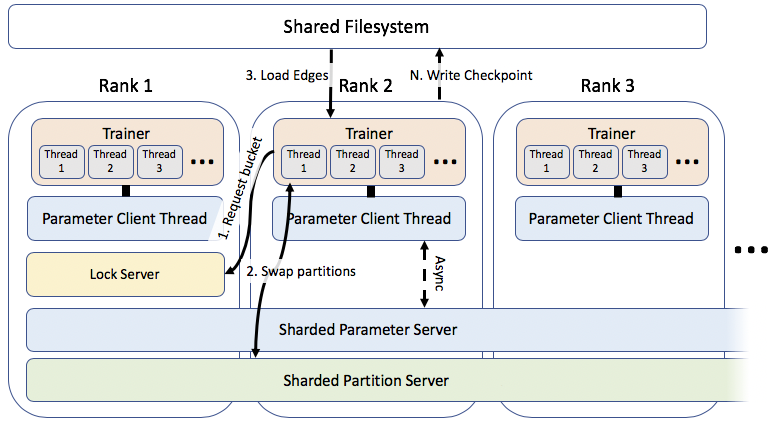}
\caption{A block diagram of the modules used for PBG's distributed mode. Arrows illustrate the communications that the Rank 2 Trainer performs for the training of one bucket. First, the trainer requests a bucket from the lock server on Rank 1, which locks that bucket's partitions. The trainer then saves any partitions that it is no longer using and loads new partitions that it needs to and from the sharded partition servers, at which point it can release its old partitions on the lock server. Edges are then loaded from a shared filesystem, and training occurs on multiple threads without inter-thread synchronization\cite{hogwild}. In a separate thread, a small number of shared parameters are continuously synchronized with a sharded parameter server. Model checkpoints are occasionally written to the shared filesystem from the trainers.}
\label{fig:distributed}
\end{figure*}

Existing distributed embedding systems typically use a parameter server architecture. In this architecture, a (possibly sharded) parameter server contains a key-value store of embeddings. At each SGD iteration, the embedding parameters required by a minibatch of data are requested from the parameter server, and gradients are (asynchronously) sent to the server to update the parameters. 

The parameter server paradigm has been effective for training large sparse models \cite{Li2014}, but it has a number of drawbacks. One issue is that parameter-server based embedding frameworks require too much network bandwidth to run efficiently, since all embeddings for each minibatch of edges and their associated negative samples must be transferred at each SGD step \cite{gridword2vec} \footnote{In fact, our approach to batched negative sampling, described in Section \ref{section:neg_sampling} reduces the number of negatives that must be retrieved so would require less bandwidth than \cite{gridword2vec} if a parameter server was used.}. Furthermore, we found it necessary for effective research use that the same models could be run in a single-machine or distributed context, but the parameter server architecture limits the size of models that can be run on a single machine. 
Finally, we would like to avoid the potential convergence problems from asynchronous model updates since our embeddings are already partitioned into independent sets.

Given partitioned entities and edges PBG employs a parallelization scheme that combines a locking scheme over the model partitions described in Section \ref{sec:partition}, with an asynchronous parameter server architecture for shared parameters i.e. the relation operators as well as unpartitioned or featurized entity types.

In this parallelization scheme, illustrated in Figure \ref{fig:distributed}, partitioned embeddings are locked by machines for training. Multiple edge buckets can be trained in parallel as long as they operate on disjoint sets of partitions, as shown in Figure \ref{fig:partitions} (left). Training can proceed in parallel on up to $P/2$ machines. The locking of partitions is handled by a centralized lock server on one machine, which parcels out buckets to the workers in order to minimize communication (i.e. favors re-using a partition) The lock server also maintains the invariant described in Section \ref{sec:partition}, that only the first bucket should operate on two uninitialized partitions.

The partitioned embeddings themselves are stored in a partition server sharded across the $N$ training machines. A machine fetches the source and destination partitions, which are often multiple GB in size, from the partition server, and trains on a bucket of edges loaded from shared disk. Checkpoints of the partitioned entities are intermittently saved to shared disk.

Some model parameters are global and thus cannot be partitioned. This most importantly includes relation parameters, as well as entity types that have very small cardinality or use featurized embeddings. There are a relatively small number of such parameters ($<10^6$), and they are handled via asynchronous updates with a sharded parameter server. Specifically, each trainer maintains a background thread that has access to all unpartitioned model parameters. This thread asynchronously fetches the parameters from the server and updates the local model, and pushes accumulated gradients from the local model to the parameter server. This thread performs continuous synchronization with some throttling to avoid saturating network bandwidth.

\subsection{Batched Negative Sampling}

\begin{figure*}[t]
\centering
\includegraphics[width=\textwidth]{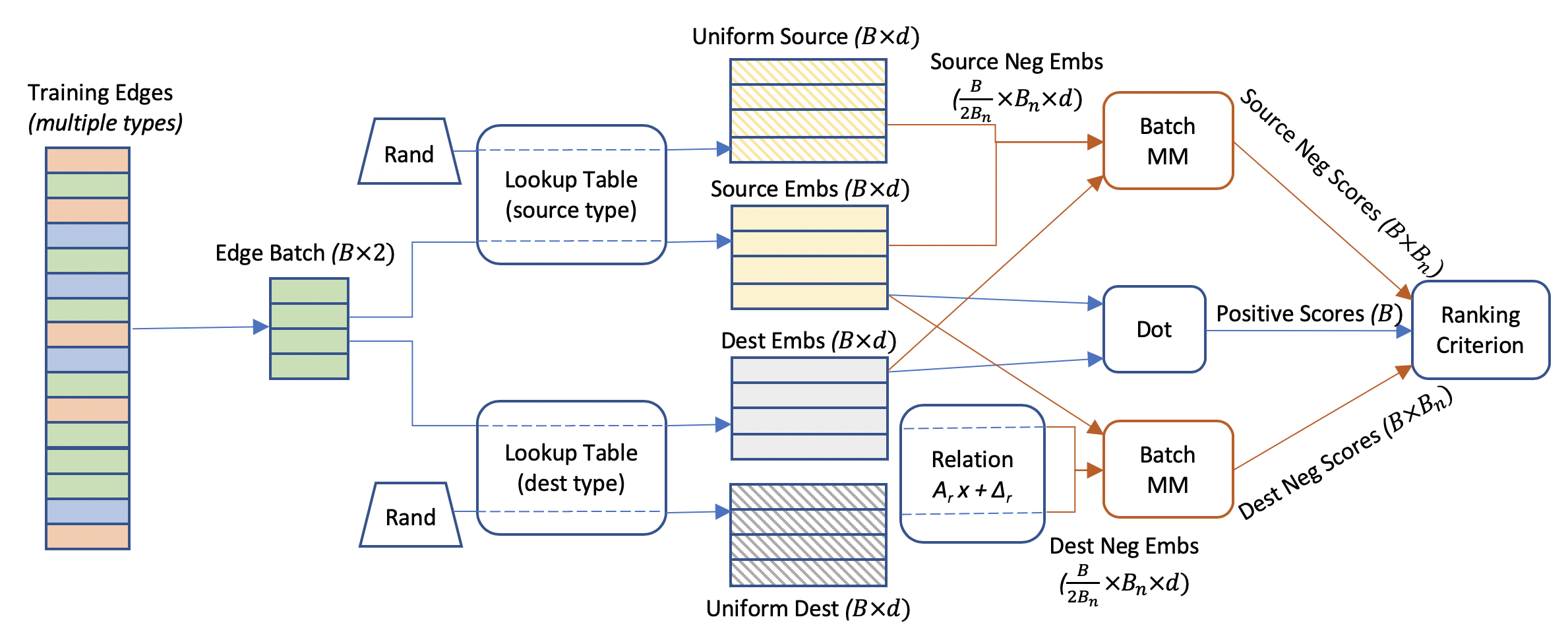}
\vspace{-5mm}

\caption{Memory-efficient batched negative sampling. Embeddings are fetched for the $B$ source and destination entities in a batch of edges, as well as $B$ uniformly-sampled source and destination entities. Each chunk of $B_n/2$ edges is corrupted with all source or destination entities in its chunk, as well as the corresponding chunk of the uniform embeddings, resulting in $B_n$ negative examples per positive edge. The negative scores are computed via a batch matrix multiply.}
\label{fig:sampling}
\end{figure*}

\label{section:neg_sampling}

\begin{figure}[h]
\centering
\includegraphics[width=0.48 \textwidth]{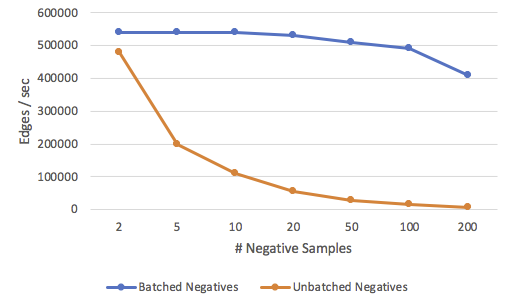}
\caption{Effect of number of negative samples per edge on training speed ($d=100$). With unbatched negatives, training speed is inversely proportional to number of negatives, but with batched negatives, speed is nearly constant for $B_n \leq 100$.}
\label{fig:batched_negs}
\end{figure}

The negative sampling approach used by most graph embedding methods is highly memory (or network) bound because it requires $B \cdot B_n \cdot d$ floats of memory access to perform $O(B \cdot B_n \cdot d$) floating-point operations ($B \cdot B_n$ dot products). Indeed, \citeauthor{wu2017starspace}. report that training speed ``is close to an inverse linear function of [number of negatives]''.

To increase memory efficiency on large graphs, we observe that a single batch of $B_n$ sampled source or destination nodes can be reused to construct multiple negative examples. In a typical setup, PBG takes a batch of $B=1000$ positive edges from the training set, and breaks it into chunks of $50$ edges. The destination (equivalently, source) embeddings from each chunk is concatenated with $50$ embeddings sampled uniformly from the tail entity type. The outer product of the $50$ positives with the $200$ sampled nodes equates to $9900$ negative examples (excluding the induced positives). The training computation is summarized in Figure \ref{fig:sampling}.

This approach is much cheaper than sampling negatives for each batch. For each batch of $B$ positive edges, only $3B$ embeddings are fetched from memory and $3 B B_n$ edge scores (dot products) are computed. The edge scores for a batch can be computed as a batched $B_n \times B_n$ matrix multiply, which can be executed with high efficiency. Figure \ref{fig:batched_negs} shows the performance of PBG with different numbers of negative samples, with and without batched negatives.

In multi-relation graphs with a small number of relations, we construct batches of edges that all share the same relation type $r$. This improves training speed specifically for the linear relation operator $f_r(t)=A_r t$, because it can be formulated as a matrix-multiply $f_r(T)=A_r T.$ 


\section{Experiments}
We evaluate PBG on two types of graphs common in both the academic literature and practical applications. 

In one set of experiments we focus on embedding real online social networks. We evaluate PBG constructed embeddings of the user-user interaction graph from LiveJournal \cite{backstrom2006group} \cite{leskovec2009community}, a user-user follow graph from Twitter \cite{kwak2010twitter} \cite{boldi2004webgraph} \cite{boldi2011layered}, and a user-user interaction graph from YouTube \cite{tang2009scalable}. The LiveJournal and Twitter data set we used are from SNAP \cite{snapnets}.

We consider two types of tasks: link prediction in the graph and the use of the graph embedding vectors to predict other attributes of the nodes. We find that PBG is much faster and more scalable than existing methods while achieving comparable performance. Second, the distributed partitioning does not impact the quality of the learned embeddings on large graphs. Third, PBG allows for parallel execution and thus can decrease wallclock training time proportional the number of partitions.

We also consider using PBG to embed the Freebase knowledge graph. Knowledge graphs have a very different structure from social networks and the presence of many relation types allows us to study the effect of using various relation operators from the literature.

Here we find that PBG can again match (or exceed) state of the art performance but that some types of relation operators (e.g. ComplEx) require care when using distributed training. 




\subsection{Experimental Setup}

\begin{table*}[t]
\centering
\begin{tabular}{l r r r r}
\hline
\multicolumn{5}{c}{\textbf{LiveJournal}} \\\hline
Metric & MRR & MR & Hits@10	& Memory\\\hline
DeepWalk* & 0.691 & 234.6 & 0.842 & 61.23 GB \\
MILE (1 level)* & 0.629 & \textbf{174.4} & 0.785 & 60.88 GB\\
MILE (5 levels)* & 0.505 & 462.8 & 0.632 & 22.78 GB
\\\hline
PBG (1 partition) & \textbf{0.749} & 245.9 & \textbf{0.857} & 20.88 GB
\\\hline 
\end{tabular}
\quad
\begin{tabular}{l r r}
\hline
\multicolumn{3}{c}{\textbf{YouTube}} \\\hline
Metric & Micro-F1 & Macro-F1 \\\hline
DeepWalk$\dagger$ & 45.2\% & 34.7\% \\
MILE (6 level)$\dagger$ & 46.1\% & 38.5\% \\
MILE (8 levels)$\dagger$ & 44.3\% & 35.3\%
\\\hline
PBG (1 partition) & \textbf{48.0}\% & \textbf{40.9}\% 
\\\hline 
\end{tabular}
\caption{\label{tab:livejournal} Performance of PBG, DeepWalk, and MILE on the LiveJournal dataset and the YouTube dataset. \textbf{Left:} Link prediction evaluation and peak memory usage for the trained embeddings of the LiveJournal dataset and YouTube dataset. The ranking metrics on the test set are obtained by ranking positive edges among randomly sampled corrupted edges. \textbf{Right:} Micro-f1 and Macro-f1 on the user categories prediction task of the YouTube dataset when using learned embeddings as features. \textbf{*} Results obtained running software provided by the original authors. $\dagger$ Results reported in \cite{liang2018mile}.
\label{tab:livejournal-result}
}
\end{table*}

For each dataset, we report the best results from a grid search of learning rates from $0.001-0.1$, margins from $0.05-0.2$ and negative batch sizes of $100-500$, and choose the parameter settings based on the validation split. Results for FB15k are reported on the separate test split. 

All experiments are performed on machines with 24 Intel\textsuperscript{\textregistered} Xeon\textsuperscript{\textregistered} cores (two sockets) and two hyperthreads per core, for a total of 48 virtual cores, and 256 GB of RAM. We use 40 HOGWILD threads for training. For distributed execution, we use a cluster of machines connected via 50Gb/s ethernet. We use the TCP backend for \texttt{torch.distributed} which in practice achieves approximately 1 GB/s send/receive bandwidth. For memory usage measurements we report peak resident set size sampled at 0.1 second intervals.




\subsection{LiveJournal}

\begin{figure}[t]
\includegraphics[width=0.48\textwidth]{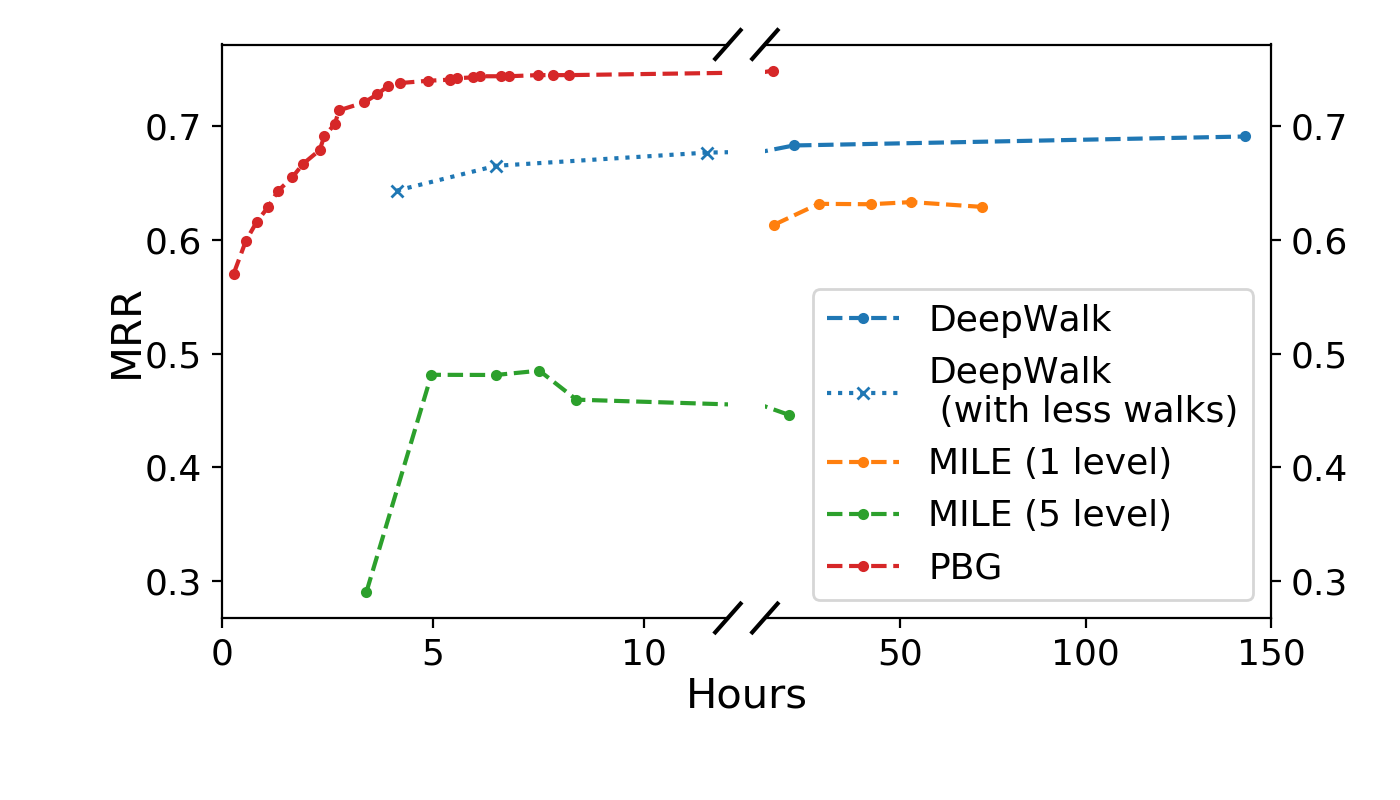}
\vspace{-6mm}
\caption{Learning curve for PBG and competing embedding methods on the LiveJournal dataset. For each approach, we evaluate the MRR after each epoch. DeepWalk takes more than 20 hours to train for a single epoch so we limit the number of walks during training for some runs to further reduce the computation time.}
\vspace{-3mm}
\label{fig:learning_curve_livejournal}
\end{figure}

We evaluate PBG performance on the LiveJournal dataset \cite{backstrom2006group,leskovec2009community} collected from the blogging site LiveJournal\footnote{https://www.livejournal.com}, where users can follow others to form a social network. The dataset contains 4,847,571 nodes and 68,993,773 edges. We construct train and test splits of the dataset that contains 75\% and 25\% of the total edges. 

We compare the PBG embedding performance with MILE, which can also scale to large graphs. MILE repeatedly coarsens the graphs into smaller ones and applies traditional embedding methods on coarsened graph at each level as well as a final refinement step to get the embeddings of the original graph. We also show the performance of DeepWalk, which is used as the base embedding method for MILE.

We evaluate the embeddings using the same methodology described in Section \ref{sec:fbfull}. To compare the computation time across different approaches, we report the learning curve of test MRR obtained by different approaches during training with respect to time (see Figure \ref{fig:learning_curve_livejournal}).

\subsection{YouTube}
To show that PBG embeddings are useful for downstream supervised tasks, we apply PBG to the Youtube dataset \cite{tang2009scalable}. The dataset contains a social network between users on YouTube\footnote{www.youtube.com}, as well as the labels of these users that represent categories of groups they subscribed. This social network dataset contains 1,138,499 nodes and 2,990,443 edges. 

We compare the performance of PBG embeddings with MILE embeddings and DeepWalk embeddings by applying those embeddings as features to perform a multi-label classification of users. We follow the typical methods \cite{deepwalk14, liang2018mile} to evaluate the embedding performance, where we run a 10-fold cross validation by randomly selecting 90\% of the labeled data as training data and the rest as testing data. We use the learned embedding as features and train a one-vs-rest logistic regression model to solve the multi-label node classfication problem. 

We find that PBG embeddings perform comparably (slightly better) than competing methods (see Table \ref{tab:livejournal-result}).







\subsection{Freebase Knowledge Graph}

\iffbone
\begin{table*}[ht]

\centering
\begin{tabular}{@{\extracolsep{4pt}}l r r r r r}
\hline
& \multicolumn{2}{c}{MRR} & \\
Method & Raw & Filtered & Hit@10 \\
\hline
RESCAL \cite{nickel2011three} & 0.189 & 0.354 & 0.587\\
TransE \cite{transE} & 0.222 & 0.463 & 0.749 \\
HolE \cite{hole} & 0.232 & 0.524 & 0.739 \\
ComplEx \cite{trouillon2016complex} & 0.242 & 0.692 & 0.840 \\
R-GCN+ \cite{schlichtkrull2018modeling} & 0.262 & 0.696 & 0.842 \\
StarSpace \cite{wu2017starspace} & - & - & 0.838 \\
Reciprocal ComplEx-N3 \cite{lacroix2018canonical} & - & 0.860 & 0.910  \\
\hline
PBG (TransE) & 0.265 & 0.594 & 0.785 \\
PBG  (ComplEx) & 0.242 & 0.790 & 0.872 \\
\hline 
\end{tabular}

\caption{\label{tab:fb15k}Comparison of PBG with other embedding methods on the FB15k dataset. PBG embeddings are trained with both a TransE and ComplEx model, and in both cases perform similarly to the reported results for that model. The best reported results on FB15k \cite{lacroix2018canonical} use extremely large embedding dimension, which we do not reproduce here.}
\label{tab:fb-result}
\end{table*}

\fi

Freebase (FB) is a large knowledge graph that contains general facts extracted from Wikipedia, etc. The FB15k dataset consists of a subset of Freebase consisting of 14,951 entities, 1345 relations and 592,213 edges. 

\iffbone
\subsubsection{FB15K}
\fi
\label{sec:fb15k}

We compare the performance of PBG embeddings on a link prediction task with existing embedding methods for knowledge graphs. We compare mean reciprocal rank and Hits@10 with existing methods for knowledge graph embeddings reported in \cite{trouillon2016complex}.\footnote{We report both raw and \textit{filtered} ranking metrics for FB15k as described in \cite{transE}. For the filtered metrics, all edges that exist in the training, validation or test sets are removed from the set of candidate corrupted edges for ranking. This avoids artificially poor results due to true edges from the data being ranked above a test edge.} Results are shown in Table \ref{tab:fb15k}.

We embed FB15k with a complex multiplication relation operator as in \cite{trouillon2016complex}. We evaluate PBG using two different configurations: one that is similar to the TransE model, and one similar to the ComplEx model.  As in that work, we also find it beneficial to use separate relation embeddings for source negatives and destination negatives (described as `reciprocal predicates' in \citealt{lacroix2018canonical}). For ComplEx, we train a 400-dimensional embedding for 50 epochs with a softmax loss over negatives using dot product similarity.

PBG performs comparably to the reported results for TransE and ComplEx models. In addition, recent papers have reported even stronger results for FB15k (and other small knowledge graphs like WordNet) using ComplEx with very large embeddings of thousands of dimensions \cite{lacroix2018canonical}. We managed to reproduce these architectures and results in the PBG framework but do not report the details here due to space constraints.

\subsubsection{Full Freebase}

\begin{table*}[ht]
\centering
\small

\begin{tabular}{l r r r r}
\hline
\# Parts & MRR & Hits@10 & Time (h) & Mem (GB) \\\hline
1  & 0.170 & 0.285 & 30  & 59.6\\
\hline
4  & 0.174 & 0.286 & 31 & 30.4\\
8  & 0.172 & 0.288 & 33 & 15.5 \\
16 & 0.174 & 0.290 & 40 & 6.8\\\hline 
\end{tabular}
\quad
\begin{tabular}{l l r r r r}
\hline
\# Machines &  \# Parts & MRR & Hits@10 & Time (h) &  Mem (GB) \\\hline
1  & 1 & 0.170 & 0.285 & 30 & 59.6 \\
\hline
2 & 4 & 0.170 & 0.280 & 23 & 64.4 \\
4 & 8 & 0.171 & 0.285 & 13 & 30.5 \\
8 & 16 & 0.163 & 0.276 & 7.7 & 15.0 \\\hline 
\end{tabular}
\caption{Model evaluation, training time (10 epochs) and peak memory usage for embeddings of the full Freebase knowledge graph. MRR and Hits@10 are evaluated in the \textit{raw} setting. \textbf{Left:} Training with different numbers of partitions on a single machine. \textbf{Right:} Distributed training on different numbers of machines. 
}
\label{tab:fb-full}
\end{table*}

\begin{figure*}[h!]
\centering
    \includegraphics[width=0.44\textwidth]{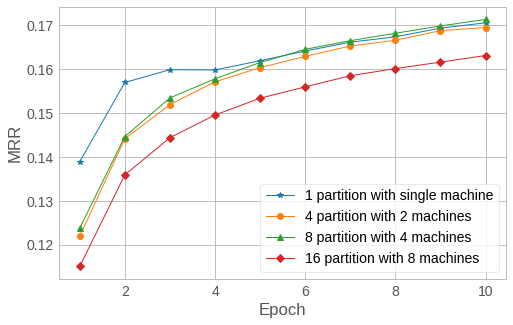}
    \includegraphics[width=0.44\textwidth]{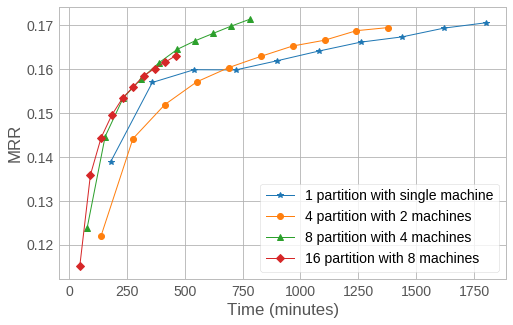}
    \vspace{-3mm}
    \caption{Learning curve for PBG models on the Freebase dataset with different number of machines used in training. MRR of learned embeddings is plotted as a function of epoch (\textbf{top}) and wallclock time (\textbf{bottom}).}
    \vspace{-3mm}
    \label{fig:freebase}
\end{figure*}

\label{sec:fbfull}
Next, we compare different numbers of partitions and distributed training using the full Freebase dataset \footnote{Google, Freebase Data Dumps, \\ https://developers.google.com/freebase, Sept. 10, 2018.} \cite{freebase:datadumps}. We use all entities and relations that appeared at least 5 times in the full dataset, resulting in a total of 121,216,723 nodes, 25,291 relations and 2,725,070,599 edges. We construct train, validation and test splits of the dataset, which contain 90\%, 5\%, 5\% of the total edges, respectively. The data format we use for the full freebase dataset is the same as in the freebase 15k dataset described in Section \ref{sec:fb15k}.

To investigate the effect of number of partitions, we partition Freebase nodes uniformly into different numbers of partitions and measure model performance, training time, and peak memory usage.  We then consider parallel training on different numbers of machines. For each number of machines $M$, we use $2M$ partitions (which is the minimum number of partitions that allows this level of parallelism. Note that the full model size ($d=100$) is 48.5 GB.

We train each model for 10 epochs, using the same grid search over hyperparameters for each number of partitions chosen from the same set grid search as FB15k. For the multi-machine evaluation, we use a consistent hyperparameters that had the best performance on single-machine training.

We evaluate the models with a link prediction task similar to that described in Section \ref{sec:fb15k}. However due to the large number of candidate nodes, for each edge in the eval set we select $10,000$ candidate negative nodes sampled from the set of entities according to their prevalence in the training data to produce negative edges which we use to compute mean reciprocal rank and hits@10\footnote{We sample candidate negative nodes according to their prevalence in the data because the full Freebase dataset has such a long-tailed degree distribution that we find that models can achieve $>50\%$ hit@1 against $10,000$ uniformly-sampled negatives, which suggests that it is just performing ranking based on the degree distribution.}.
We report these results raw (unfiltered), following prior work on large graphs \cite{transE}. 

Results are reported in Table \ref{tab:fb-full}, along with training time and memory usage.

We observe that on a single machine, peak memory usage decreases almost linearly with number of partitions, but training time increases somewhat due to extra time spent on I/O\footnote{This I/O overhead is higher on sparser graphs and lower on denser graphs.}. On multiple machines, the full model is sharded across the machines rather than on disk, so memory usage is higher with 2 machines, but decreases linearly as the number of machines increases. Training time also decreases with increasing number of machines, although there is again some overhead for training on multiple machines. This consists of a combination of I/O overhead and incomplete occupancy. The occupancy issue arises because there may not always be an available bucket with non-locked partitions for a machine to work on. Increasing the number of partitions relative to the number of machines will thus increase occupancy, but we don't examine this tradeoff in detail.

Freebase embeddings have nearly identical link prediction accuracy after 10 epochs of training with and without node partitioning and parallelization up to four machines.  For the highest parallelization condition (8 machines), a small degradation in MRR from $0.171$ to $0.163$ is observed.

PBG embeddings trained with the ComplEx model perform better than TransE on the link prediction task, achieving MRR of $0.24$ and Hits@10 of $0.32$ with $d=200$ and a single partition. However, our experiments show that training ComplEx models with multiple partitions and machines is unstable, and MRR varies from $0.15$ to $0.22$ across replicates. Further investigation of the performance of ComplEx models via PBG partitioning is left for future work. 





\subsection{Twitter} 

\begin{table*}[t]
\centering
\small

\begin{tabular}{l r r r r}
\hline
\# Parts & MRR & Hits@10 & Time (h) & Mem (GB) \\\hline
1  & 0.136 & 0.233 & 18.0 & 95.1 \\
\hline
4  & 0.137 & 0.235 & 16.8 & 43.4 \\
8  & 0.137 & 0.237 & 19.1 & 20.7 \\
16 & 0.136 & 0.235 & 23.8 & 10.2 \\\hline 
\end{tabular}
\quad
\begin{tabular}{l l r r r r}
\hline
\# Machines &  \# Parts & MRR & Hits@10 & Time (h) &  Mem (GB) \\\hline
1  & 1 & 0.136 & 0.233 & 18.0 & 95.1 \\
\hline
2 & 4 & 0.137 & 0.235 & 9.8 & 79.4 \\
4 & 8 & 0.137 & 0.235 & 6.5 & 40.5 \\
8 & 16 & 0.137 & 0.235 & 3.4 & 20.4 \\\hline 
\end{tabular}
\caption{Model evaluation, training time (10 epochs) and peak memory usage for embeddings of the Twitter graph. \textbf{Left:} Training with different numbers of partitions on a single machine. \textbf{Right:} Distributed training on different numbers of machines. 
}
\label{tab:twitter}
\end{table*}

\begin{figure*}[h]
\centering
    \includegraphics[width=0.44\textwidth]{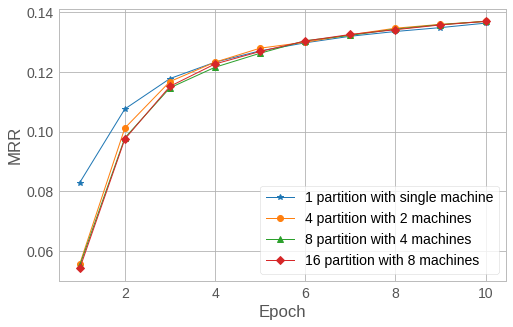}
    \includegraphics[width=0.44\textwidth]{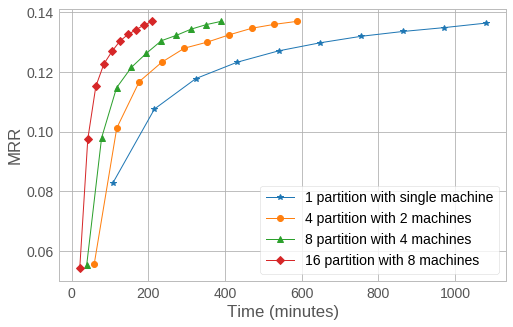}
    \vspace{-3mm}
    \caption{Learning curve for PBG models on the Twitter dataset with different number of machines used in training. MRR of learned embeddings is plotted as a function of epoch (\textbf{top}) and wallclock time (\textbf{bottom}).}
    \vspace{-3mm}
    \label{fig:twitter}
\end{figure*}

Finally, we consider the scaling of PBG on a social network graph in comparison to the Freebase knowledge graph studied in Section \ref{sec:fbfull}. We embed a publicly available Twitter \footnote{www.twitter.com} subgraph \cite{kwak2010twitter} \cite{boldi2004webgraph} \cite{boldi2011layered} containing a social network between 41,652,230 nodes and 1,468,365,182 edges with a single relation called ``follow''. We construct train, validation and test splits of the dataset, which contain 90\%, 5\%, 5\% of the total edges, respectively.

In Table \ref{tab:twitter} we report MRR and Hits@10 after 10 training epochs as well as training time and peak memory usage for different partitioning and parallelization schemes. The results are consistent with Table \ref{tab:fb-full}: we observe a decrease in training time with multiple machines, without a loss in link prediction accuracy up to 8 machines.

In Figure \ref{fig:twitter} we report the learning curve of test MRR obtained by different number of machines used during training with respect to epoch and time. Compared to the Freebase knowledge base learning curves in Figure \ref{fig:freebase}, the Twitter graph shows more linear scaling of training time as the graph is partitioned and trained in parallel.

\section{Conclusion}
In this paper, we present PyTorch-BigGraph, an embedding system that scales to graphs with billions of nodes and trillions of edges. PBG supports multi-entity, multi-relation graphs with per-relation configuration such as edge weight and choice of relation operator. To save on memory usage and to allow parallelization PBG performs a block decomposition of the adjacency matrix into $N$ buckets, training on the edges from one bucket at a time. 

We show that the quality of embeddings trained with PBG are comparable with existing embedding systems, and require less time to train. We show that partitioning of the Freebase graph reduces memory consumption by 88\% without degrading embedding quality, and distributed execution on 8 machines speeds up training by a factor of 4. Our experiments have shown that embedding quality is quite robust to partitioning and parallelization in social network datasets, but may be more sensitive to parallelization when the number of relations is large, the degree distribution is highly skewed, or relation operators such as ComplEx are used. Thus improving the scaling for these more complicated models is an important area for future research.



We have presented PBG's performance on the largest publicly available graph datasets that we are aware of. However, the largest benefits of the PBG architecture come from graphs that are $1-2$ orders of magnitude larger than these, where more fine-grained partitioning is necessary and exposes more parallelism. We hope that this work and the open source release of PBG helps to motivate the release of larger graph datasets and an increase in research and reported results on larger graphs.

\section{Acknowledgements}

We would like to acknowledge Adam Fisch, Keith Adams, Jason Weston, Antoine Bordes and Serkan Piantino for helping to formulate the initial ideas that led to this work, as well as Maximilian Nickel who provided helpful feedback on the manuscript.
\section{Reference}

\bibliographystyle{sysml2019}
\bibliography{reference}

\end{document}